\documentclass{article}

\usepackage[preprint]{neurips_2025}

\usepackage[utf8]{inputenc} % allow utf-8 input
\usepackage[T1]{fontenc}    % use 8-bit T1 fonts
\usepackage{hyperref}       % hyperlinks
\usepackage{url}            % simple URL typesetting
\usepackage{booktabs}       % professional-quality tables
\usepackage{amsfonts}       % blackboard math symbols
\usepackage{amsmath}    % for advanced math environments
\usepackage{amssymb}
\usepackage{nicefrac}       % compact symbols for 1/2, etc.
\usepackage{microtype}      % microtypography
\usepackage{xcolor}         % colors
\usepackage{graphicx}

\title{Towards fully differentiable neural ocean model with Veros}

\author{%
  Etienne Meunier\textsuperscript{1} \hspace{0.8em}
  Said Ouala\textsuperscript{2} \hspace{0.8em}
  Hugo Frezat\textsuperscript{3} \hspace{0.8em}
  Julien Le Sommer\textsuperscript{3} \hspace{0.8em}
  Ronan Fablet\textsuperscript{2} \\
  \textsuperscript{1}Inria Paris, France \quad
  \textsuperscript{2}IMT Atlantique, France \quad
  \textsuperscript{3}IGE, Grenoble, France
}

\begin{document}

\maketitle

\begin{abstract}
We present a differentiable extension of the VEROS ocean model \cite{veros}, enabling automatic differentiation through its dynamical core. We describe the key modifications required to make the model fully compatible with JAX’s autodifferentiation framework and evaluate the numerical consistency of the resulting implementation. Two illustrative applications are then demonstrated: (i) the correction of an initial ocean state through gradient-based optimization, and (ii) the calibration of unknown physical parameters directly from model observations. These examples highlight how differentiable programming can facilitate end-to-end learning and parameter tuning in ocean modeling. Our implementation is available online .\footnote{https://github.com/Etienne-Meunier/Veros-Autodiff}
\end{abstract}

\section{Introduction}
Earth system models (ESMs) are widely used to study climate changes resulting from both anthropogenic and natural perturbations. Over the past years, significant advances have been made through the development of new numerical schemes, refined physical parameterizations, and the use of increasingly powerful computers. Despite these advances, tuning ESMs to accurately reproduce historical data remains largely a manual process \cite{hourdin2017art}, and persistent errors and biases continue to challenge their accuracy.

Recent advances in scientific machine learning motivated the development of new learning-based methods for the design and calibration of ESMs \cite{irrgang2021towards,nguyen2023climax,ouala2024online,dheeshjith2025samudra}. In this context, one promising way of building climate models that are informed with Machine Learning (ML) is based on Differentiable Programing (DP) paradigms \cite{gelbrecht2023differentiable,sapienza2024differentiable}. In particular, DP methods allow for designing calibration methods that can be trained end-to-end \cite{frezat2022posteriori,kochkov2023neural} which, in turn, has the potential to unify and solve various challenges in climate modeling including, model bias correction, parameter tuning and the design of subgrid-scale parameterizations that are informed by observations. However, most existing climate models are implemented in programming languages that do not natively support automatic differentiation. Moreover, they are often based on design choices aimed towards memory efficiency (e.g., in-place operations), which prevents the use of automatic differentiation methods.% such as backpropagation.

Despite these limitations, recent studies have explored the development of dynamical cores that natively support automatic differentiation and can be trained end-to-end \cite[e.g.][for atmospheric simulations]{kochkov2023neural}. In this work, we take a step in this direction by exploring ocean model calibration within a differentiable programming framework. Specifically, we use the VERsatile Ocean Simulator (VEROS) \cite{veros}, which contains a JAX backend, to evaluate the ability of using DP to ocean model tuning. We show how we adapted VEROS to support JAX's automatic differentiation and present examples demonstrating the use of both forward and reverse mode differentiation for state and parameter calibration.

\section{Differentiability properties in Veros}
\subsection{Modifications to Veros}

To enable differentiation through the Veros step function, we implemented several key modifications to ensure compatibility with JAX's automatic differentiation framework.

\paragraph{Functionally pure step function.}
The original Veros step function employs in-place operations that directly modify state variables, which violates the functional purity requirements of JAX automatic differentiation. To address this incompatibility, we developed a wrapper function that creates a copy of the input state and returns the updated state without modifying the original input. This approach ensures functional purity for both forward- and reverse-mode gradient computations. We also implemented a selective differentiation wrapper that freezes all variables except for a specified subset, limiting gradient computation to only the parameters of interest.

\paragraph{Handling functions with singular gradients.}
While most operations in Veros are inherently differentiable, certain functions exhibit undefined or singular gradients that can lead to numerical instabilities. For example, the square root function $f(x) = \sqrt{x}$ has an undefined gradient at $x = 0$. To address this, we preserve the original forward computation but define custom backward passes. Specifically, for the square root, we implement the regularized gradient
$f'(x) \triangleq (2 \sqrt{\max(x, \varepsilon)})^{-1}$ where $\varepsilon > 0$ is a small regularization parameter that prevents singularities at the origin. JAX's custom gradient functionality enables this selective modification of backward passes.

\paragraph{Computational graph preservation.}
We eliminated in-place type conversions that could break the computational graph required for automatic differentiation. These modifications primarily affected diagnostic routines. Additionally, we replaced standard Python \texttt{assert} statements with JAX's \texttt{checkify} framework\footnote{\url{https://docs.jax.dev/en/latest/debugging/checkify_guide.html}} to maintain compatibility with JAX transformations.

\subsection{Forward behaviour}

We rigorously verified that our modifications preserve the original forward simulation behavior. All control flow structures and operations were replaced with JAX-compatible equivalents where necessary, and when adjustments were required, only gradient definitions were modified while preserving forward computations unchanged.  To validate forward compatibility, we compared temperature fields after 3000 integration steps between our modified implementation and the original Veros code. The relative error was approximately $10^{-10}$, which we attribute to floating-point precision differences between backends. The computational performance of forward simulation remains virtually unchanged, as the core numerical operations are identical to the original implementation.

\subsection{Backward behaviour}%Gradient accuracy verification}

We validated the correctness of computed gradients through comparison with finite-difference approximations. For a composite function $\ell \triangleq g(s(x)) : \mathbb{R}^{\mathcal{S}} \to \mathbb{R}$, where $s$ represents the Veros step function and $g$ is a scalar aggregation operator, we compute the gradient validation error:

\[
\begin{alignedat}{2}
    \mathcal{E} &= \biggl\lVert \nabla_x \ell \big|_{x = w} \cdot \mathbf{k}
    - \frac{\ell(w + \varepsilon \mathbf{k}) - \ell(w - \varepsilon \mathbf{k})}{2\varepsilon}
    \biggr\rVert_2
    &\qquad
    \ell &\triangleq g(s(x)) : \mathbb{R}^{\mathcal S} \to \mathbb{R}
\end{alignedat}
\]

Here, $\nabla_x \ell$ is computed using automatic differentiation, $\mathbf{k}$ is a random unit vector, $w$ is the evaluation point, and $\varepsilon \approx 10^{-4}$ is the finite-difference step size. Over single-step Veros evaluations, this validation yields errors on the order of $10^{-7}$, confirming the accuracy of our gradient implementation. Additional details regarding accuracy across multiple time steps are provided in the Appendix.

\subsection{Toy experiment: gradient-based correction of initial states}
\label{toy}

To validate our gradient implementation and demonstrates practical utility for data assimilation, we design an inverse problem that reconstructs an original temperature field from a perturbed initial condition through optimization across multiple Veros steps.

\textbf{Setup :} Starting from a reference initial temperature field $T_{\text{ref}}(t=0)$, we create a perturbed version by adding a spatial Gaussian perturbation centered in the domain. We then minimize the L2 loss between temperature fields after $l=4$ integration steps: $\mathcal{L}(T_0) = \|S^{(4)}(T_0) - S^{(4)}(T_{\text{ref}})\|_2^2$
where $S$ represents Veros integration step and $T_0$ is the optimized initial field. Gradient descent is used: $T_0^{(k+1)} = T_0^{(k)} - \alpha \nabla_{T_0} \mathcal{L}(T_0^{(k)})$.

\begin{figure}
    \centering
    \includegraphics[width=\linewidth]{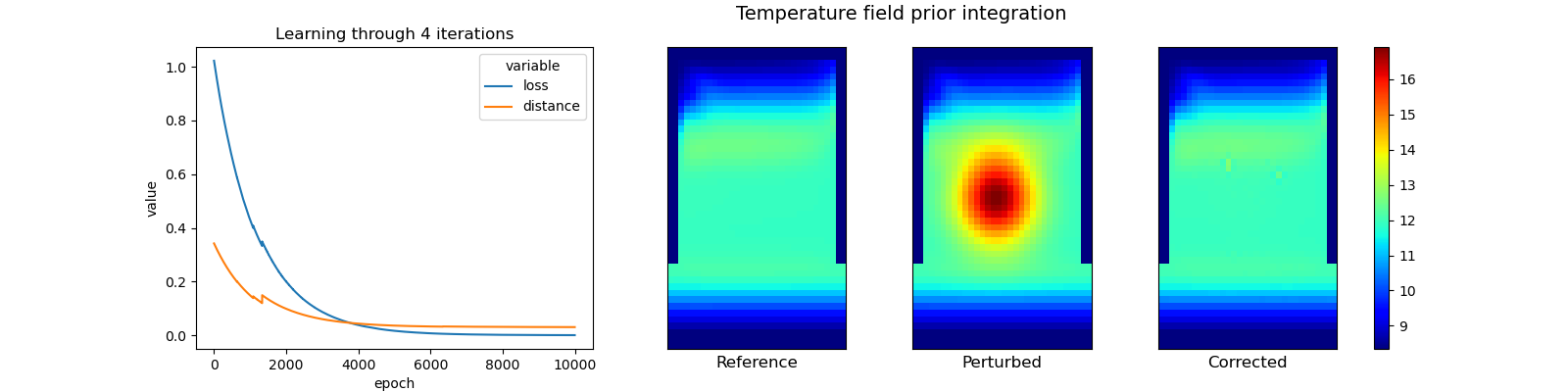}
    \caption{Results from the initial field reconstruction experiment. Left: Loss and distance metrics during optimization. Right: Temperature field snapshots showing the reference field, initial perturbed field, and final reconstructed field after optimization.}
    \label{fig:learning}
\end{figure}

\textbf{Results :} Figure \ref{fig:learning} shows a successful field reconstruction. The left panel displays the loss function ($\mathcal{L}(T_0)$) and distance metric ($\|T_0^{(k)} - T_{\text{ref}}\|_2^2$) during optimization—both converge to near-zero. The right panel shows temperature snapshots: reference, perturbed initial, and final reconstructed fields, visually confirming successful recovery of the original distribution.

%This experiment validates both gradient accuracy through multiple Veros steps and demonstrates practical utility for inverse problems relevant to data assimilation applications.

\section{Application to model calibration problems
}

\subsection{Idealised Antarctic Circumpolar Current (ACC) model configuration}

We use the ACC model configuration from the VEROS setup gallery at an eddy-permitting resolution of 1/4°. The configuration consists of a zonally re-entrant channel representing an idealized ACC, connected to a northern ocean basin enclosed by land, representing the Atlantic ocean. The model domain extends from 0° to 60°E and from 40°S to 40°N, with 15 vertical levels resolving the water column down to 2000 m depth. The circulation is forced by a meridionally-varying eastward wind stress over the ACC region, combined with a surface temperature relaxation. More details on simulation parameters are provided in the appendix. 

This simulation setup is used to demonstrate the potential of automatic differentiation for parameter calibration in ocean models. We focus on the lateral viscosity ($A_h$) and bottom friction coefficients ($r_{\text{bot}}$), as these parameters strongly influence resolved mesoscale variability. Bottom friction is a primary energy sink for mesoscale kinetic energy, which alters the vertical structure of resolved eddies, and can change their amplitude and scales \cite{Arbic2008}. Lateral viscosity controls small-scale dissipation and numerical stability, therefore affecting eddy lifetimes, spectral slopes and the cascade of energy/enstrophy \cite{Pearson2017}. The calibration of these two parameters provide a framework to mimic calibration experiments based on satellite observations.

A reference simulation is first generated using prescribed values for these parameters (see Tab.~\ref{tab:params} for details). We then demonstrate that automatic differentiation enables the recovery of these reference values when starting from mis-specified parameter choices, relying solely on observations of the barotropic streamfunction (BSF), which, in this idealized setup, serves as a proxy for large-scale circulation features that are typically inferred from satellite observations of sea surface height (SSH).

\subsection{Sensitivity analysis for parameter calibration}
\begin{figure}[ht]
    \centering
    \includegraphics[height=3.4cm]{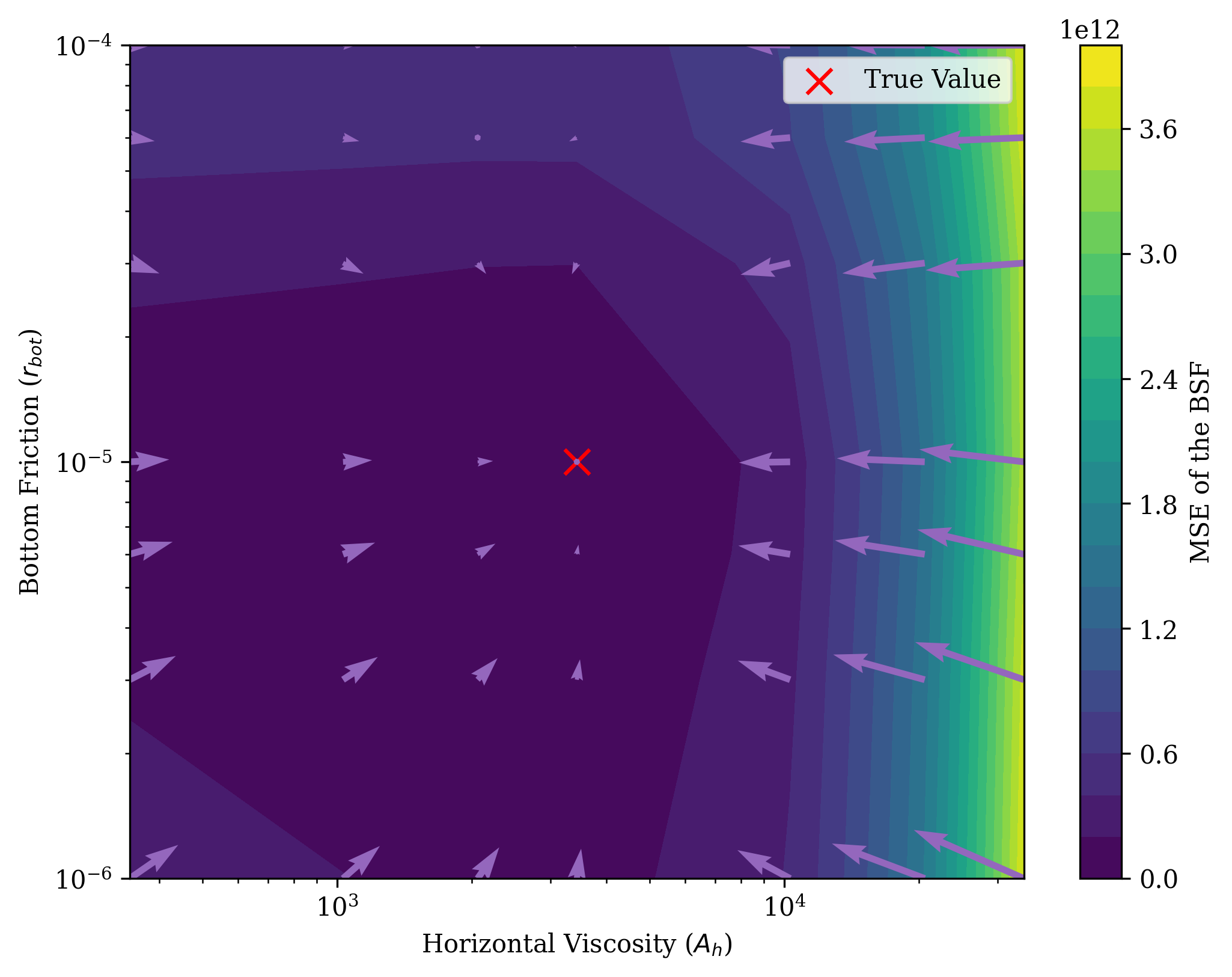}\vrule\vrule\vrule\includegraphics[height=3.5cm]{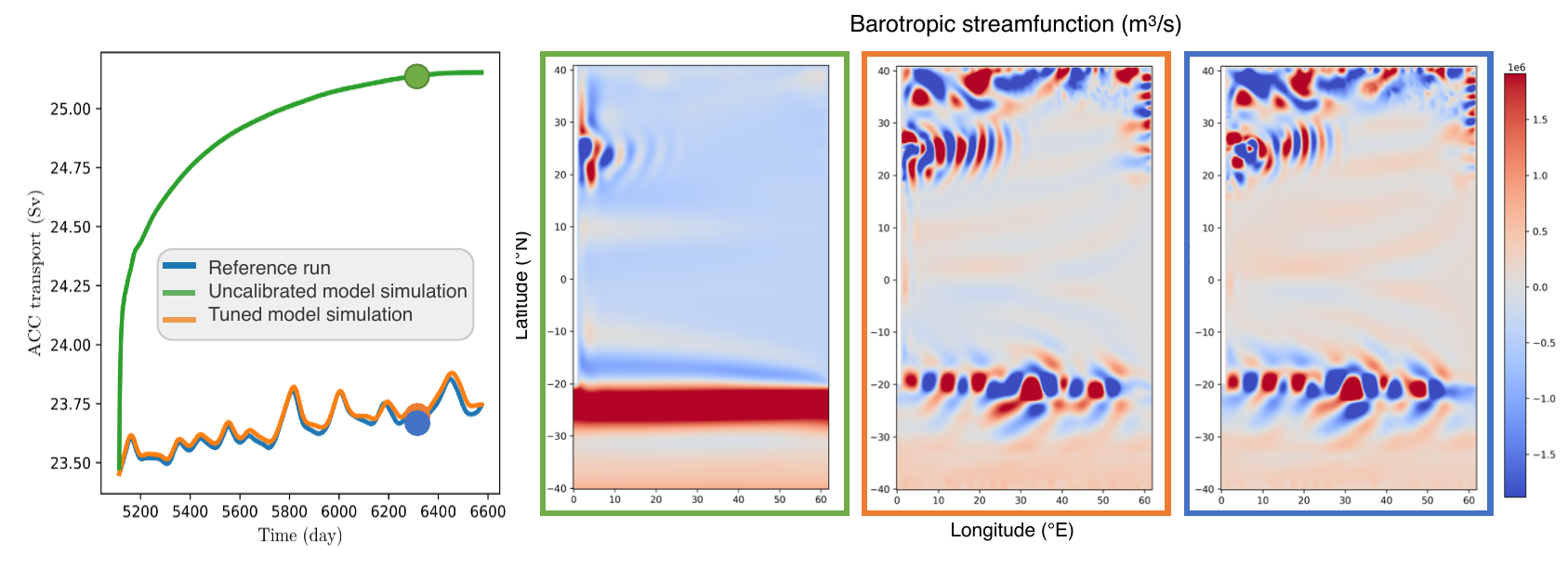}
    \caption{
        \textit{Left:} Loss landscape for the calibration task. Color indicates the loss for each combination of parameter and vector field gradients direction. 
        \textit{Right:} Results of calibration experiment.  ACC transport (Sverdrup) and Snapshots of the barotropic streamfunction.
    }
    \label{fig:side_by_side}
\end{figure}

The left panel of Fig.~\ref{fig:side_by_side} shows the gradient of the loss function with respect to both parameters, computed using forward-mode automatic differentiation. The loss is defined as the mean squared error between the simulated and reference BSF. The gradient field is visualized using a quiver plot, where arrows indicate the direction and magnitude of the steepest descent in parameter space. Overall, the gradient magnitudes differ significantly between the two parameters. The loss function exhibits stronger sensitivity to variations in $A_h$ (horizontal direction) compared to $r_{\text{bot}}$ (vertical direction), as shown by the predominance of gradient vectors in the horizontal orientation for most regions of the parameter space. This anisotropy suggests that lateral viscosity has a more direct influence on the large-scale circulation patterns captured by the BSF. Beyond this scale difference in gradient magnitudes, the direction of the gradient vectors generally points toward a convergence region that is consistent with the location of the true parameter values (marked by red crosses), which suggests that gradient-based optimization is suitable for recovering the true values of these parameters. %However, the presence of regions with weak gradients and the scale difference between the gradients of the two parameters indicates that using adaptive learning rates per variable may be necessary for efficient calibration.

% \begin{figure}[ht]
%     \centering
%     % --- Left figure ---
%     \begin{minipage}[t]{height=4cm}
%         \centering
%         \includegraphics[width=\linewidth, keepaspectratio]{figures/gradient_landscape.png}
%         \caption{Loss landscape for the calibration task. Color indicates the loss for each combination of parameter and vector field gradients direction.}
%         \label{fig:gradient_landscape}
%     \end{minipage}
%     \hfill
%     % --- Right figure ---
%     \begin{minipage}[t]{height=4cm}}
%         \centering
%         \includegraphics[width=\linewidth, keepaspectratio]{figures/figureaccveros.pdf}
%         \caption{Results of calibration experiment. Left: ACC transport (Sverdrup) for the reference run, uncalibrated simulation, and tuned simulation. Right: Snapshot of the barotropic streamfunction showing physical similarity between the tuned and reference simulations, in contrast to the uncalibrated one.}
%         \label{fig:calibration_analysis}
%     \end{minipage}
% \end{figure}

\subsection{Calibration experiment analysis}
In this experiment, we evaluate the ability of using gradient-based optimization to tune model parameters using observations of the BSF. As highlighted in right panel of Fig.~\ref{fig:side_by_side}, the uncalibrated model uses values of $A_h$ and $r_{\text{bot}}$ that result in quantitative changes in the ACC transport and in the distribution of mesoscale eddies, illustrated by the BSF anomaly relative to the long-term mean.

Overall, the gradient-based calibration yields parameter values that are very close to the reference values, and the resulting calibrated simulation closely reproduces the reference run, both in terms of ACC transport strength and the spatial distribution of mesoscale eddies.

\section{Conclusion and future work}

In this work, we introduce modifications to the Veros ocean simulator to enable automatic differentiation capabilities. We demonstrate gradient computation through two examples: correcting an initial state and calibrating physical parameters based on a reference simulation. This provides essential tools for online learning of parameterizations and model correction in ocean modeling.

Our implementation successfully computes accurate gradients through the complex nonlinear dynamics of Veros in the demonstrated scenarios, opening possibilities for data assimilation, parameter estimation, and physics-informed machine learning in oceanography. The validation experiments confirm gradient accuracy and practical utility, though further investigation is needed to assess performance over extended simulation horizons.

Future work will focus on optimizing gradient computation performance, particularly memory usage and processing time for long rollouts. Key directions include implementing implicit differentiation schemes \cite{blondel2022efficient} through iterative solvers in vertical mixing schemes, which currently represent computational bottlenecks. We also plan to explore hybrid approaches combining automatic differentiation with adjoint sensitivity methods \cite{pontryagin2018mathematical} for improved efficiency in long-horizon problems, while conducting experiments to validate gradient accuracy over extended temporal scales.

\acksection

This study is a contribution to AI4PEX (Artificial Intelligence and Machine Learning for Enhanced Representation of Processes and Extremes in Earth System Models), a project supported by the European Union’s Horizon Europe research and innovation programme under grant agreement no. 101137682. This study also contributes to EDITO Model Lab, a project supported by the European Union’s Horizon Europe research and innovation programme under grant agreement no. 101093293. 

This study has received funding from Agence Nationale de la Recherche - France 2030 as part of the PEPR TRACCS programme under grant number ANR-22-EXTR-0006 and ANR-22-EXTR-0007.

\bibliographystyle{plainnat} % or abbrvnat, unsrtnat, etc., as required
\bibliography{biblio}

\section*{Appendix}

\subsection{Experimental configuration}

Our experiments utilize Veros with the JAX backend and the \texttt{scipy\_jax} solver for linear systems. We enable the energetically consistent kinetic energy (EKE) parameterization while disabling both the turbulent kinetic energy (TKE) and neutral diffusion parameterizations to focus on the core differentiable dynamics.

\subsection{Gradient validation}

To assess the effect of repeated time-stepping on gradient accuracy, we compute $\tfrac{\partial \mathcal{L}}{\partial r_{\text{bot}}}$, where $r_{\text{bot}}$ is the bottom drag coefficient tuned in our first experiment. As shown in Fig.~\ref{fig:gradients_acc}, the discrepancy between gradient estimation methods increases with the number of time steps.

\begin{figure}[!h]
    \centering
    \includegraphics[width=\linewidth]{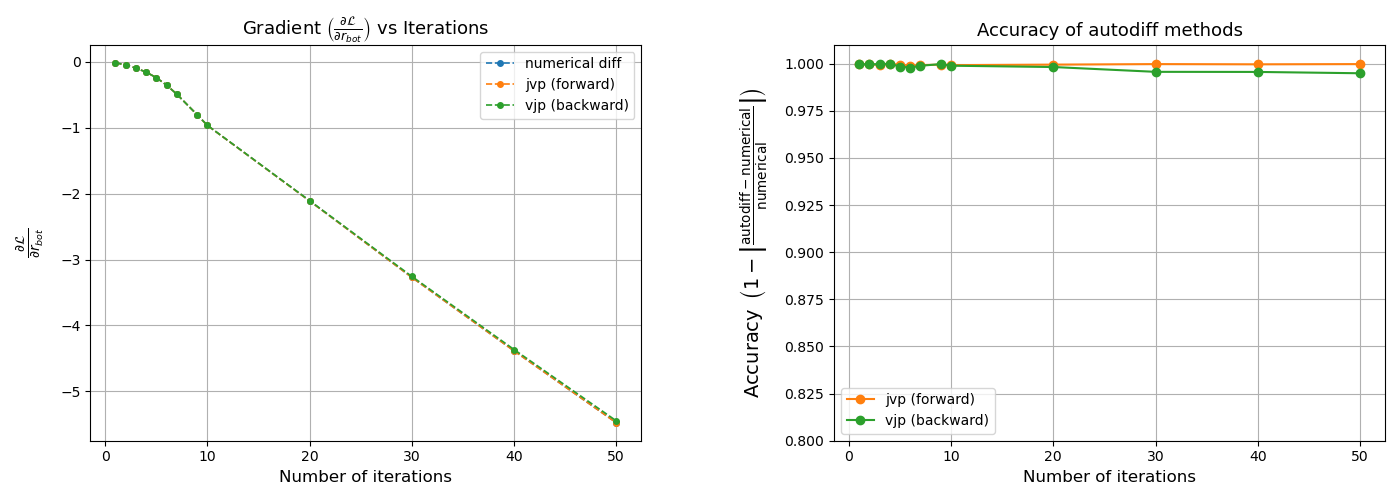}
    \caption{Gradient of the loss with respect to $r_{\rm bot}$ and the accuracy of automatic differentiation methods over iterations. 
    \textbf{Left:} Evolution of $\partial \mathcal{L} / \partial r_{\rm bot}$ for different methods. 
    \textbf{Right:} Accuracy of forward-mode (JVP) and reverse-mode (VJP) automatic differentiation, computed as $1 - |\text{autodiff grad} - \text{finite diff}| / |\text{finite diff}|$, highlighting the agreement with numerical gradients over iterations.}
    
    \label{fig:gradients_acc}
\end{figure}

\subsection{Gradient computational cost}

We analyze the computational overhead of gradient computation by measuring the cost of vector-Jacobian products (VJP) through Veros integration steps using the experimental setup from Section \ref{toy}. Figure \ref{fig:gradients_acc} demonstrates that gradient computation cost scales linearly with the number of integration steps.

\begin{figure}[!h]
    \centering
    \includegraphics[width=0.5\linewidth]{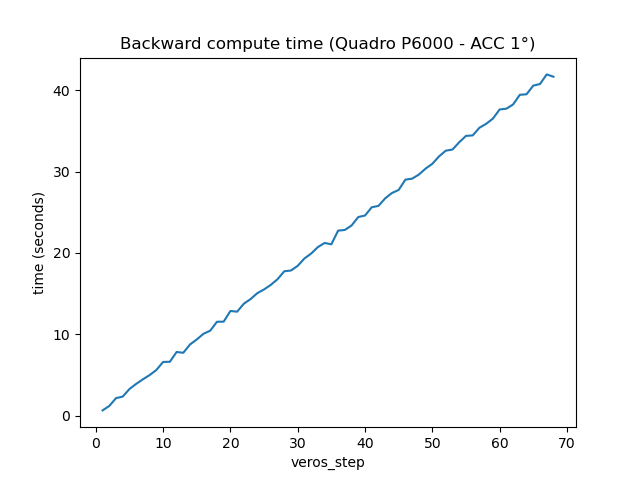}
    \caption{Gradient computation time using vector-Jacobian products (VJP) for optimizing the initial temperature field with respect to a loss function evaluated after multiple Veros integration steps.}
    \label{fig:gradients_acc}
\end{figure}

\subsection{ACC model configuration}

\begin{table}[htbp]
\centering
\caption{Model configuration parameters for the idealized ACC simulation. Parameters marked with * are the calibration targets in our experiments. For detailed descriptions of these variables, refer to the VEROS documentation.}
\label{tab:params}
\begin{tabular}{ll}
\hline
\textbf{Parameter} & \textbf{Value} \\
\hline
\multicolumn{2}{l}{\textit{Domain configuration}} \\
Resolution & 1/4° \\
Domain extent & 0°--60°E, 40°S--40°N \\
Vertical levels & 15 \\
Maximum depth & 2000 m \\
\hline
\multicolumn{2}{l}{\textit{Time stepping}} \\
$\Delta t_{\text{mom}}$ & 1200 s \\
$\Delta t_{\text{tracer}}$ & 1200 s \\
Run length & 4 years \\
\hline
\multicolumn{2}{l}{\textit{Friction and mixing parameters}} \\
$A_h^*$ & $3435.5036038313715$ m$^2$/s \\
$r_{\text{bot}}^*$ & $10^{-5}$ s$^{-1}$ \\
$K_{\text{iso}}$ & 62.5 m$^2$/s \\
$K_{\text{GM}}$ & 62.5 m$^2$/s \\
\hline
\multicolumn{2}{l}{\textit{Eddy kinetic energy (EKE)}} \\
$c_k$ (EKE) & 0.4 \\
$c_\varepsilon$ (EKE) & 0.5 \\
EKE$_{\text{max}}$ & 625 m$^2$/s$^2$ \\
$L_{\text{min}}$ & 1 m \\
\hline
\multicolumn{2}{l}{\textit{Turbulent kinetic energy (TKE)}} \\
$c_k$ (TKE) & 0.1 \\
$c_\varepsilon$ (TKE) & 0.7 \\
$\alpha_{\text{TKE}}$ & 30.0 \\
$\kappa_M^{\text{min}}$ & $2 \times 10^{-4}$ m$^2$/s \\
$\kappa_H^{\text{min}}$ & $2 \times 10^{-5}$ m$^2$/s \\
\hline
\end{tabular}
\end{table}
\end{document}